\pdfoutput=1
\documentclass[runningheads]{llncs}
\usepackage[T1]{fontenc}
\usepackage{graphicx}
\usepackage{booktabs}%
\usepackage{listings}%
\usepackage{subcaption}
\usepackage{amsmath}
\usepackage{amsfonts}
\usepackage{float}
\begin{document}
\title{Look Within, Why LLMs Hallucinate: A Causal Perspective}
%
%
\author{He Li\inst{1}\orcidID{0009-0006-5709-4636} \and
Haoang Chi\inst{1}\orcidID{0000-0002-2644-3323} \and
Mingyu Liu\inst{1}\orcidID{0009-0006-8628-0614} \and
Wengjing Yang\inst{1}\orcidID{0000-0002-6997-0406}
} 
\authorrunning{H. Li et al.}
%
\institute{
     National University of Defense Technology, Changsha, China 
}

\maketitle              
\begin{abstract}
The emergence of large language models (LLMs) is a milestone in generative artificial intelligence, achieving significant success in text comprehension and generation tasks.
Despite the tremendous success of LLMs in many downstream tasks, they suffer from severe hallucination problems, posing significant challenges to the practical applications of LLMs. Most of the works about LLMs' hallucinations focus on data quality.
Self-attention is a core module in transformer-based LLMs, while its potential relationship with LLMs' hallucination has been hardly investigated.
To fill this gap, we study this problem from a causal perspective.
We propose a method to intervene in LLMs' self-attention layers and maintain their structures and sizes intact.
Specifically, we disable different self-attention layers in several popular open-source LLMs and then compare their degrees of hallucination with the original ones. 
We evaluate the intervened LLMs on hallucination assessment benchmarks and conclude that disabling some specific self-attention layers in the front or tail of the LLMs can alleviate hallucination issues. The study paves a new way for understanding and mitigating LLMs' hallucinations.

\keywords{Hallucination \and Large Language Models \and Causal Representation Learning}
\end{abstract}
\section{Introduction}\label{sec1}
The emergence of large language models (LLMs) marks a milestone in generative artificial intelligence and natural language processing (NLP) \cite{huang2023survey,zhang2023siren}.
LLMs are typically pre-trained on extensive unlabeled corpora and fine-tuned on specific task datasets.
LLMs, such as LLaMA \cite{touvron2023LLaMA2}, LLaMA-2 \cite{touvron2023LLaMA}, Gemma \cite{team2024gemma}, Mistral \cite{jiang2023mistral}, GPT-3 \cite{brown2020language}, PaLM \cite{chowdhery2023palm}, BLOOM \cite{workshop2022bloom}, GPT-4 \cite{openai2023gpt}, have achieved tremendous success in tasks such as text generation, text comprehension \cite{chia2023instructeval}, logical reasoning \cite{hendrycks2020measuring,hendrycks2021measuring}, and have demonstrated excellent in-context learning abilities \cite{dai2022can,liu2021makes}.
This powerful capability gives LLMs tremendous practical value in real-world scenarios, such as assisting in healthcare and coding. 

\begin{figure}[t]
    \centering
    \includegraphics[width=0.8\textwidth]{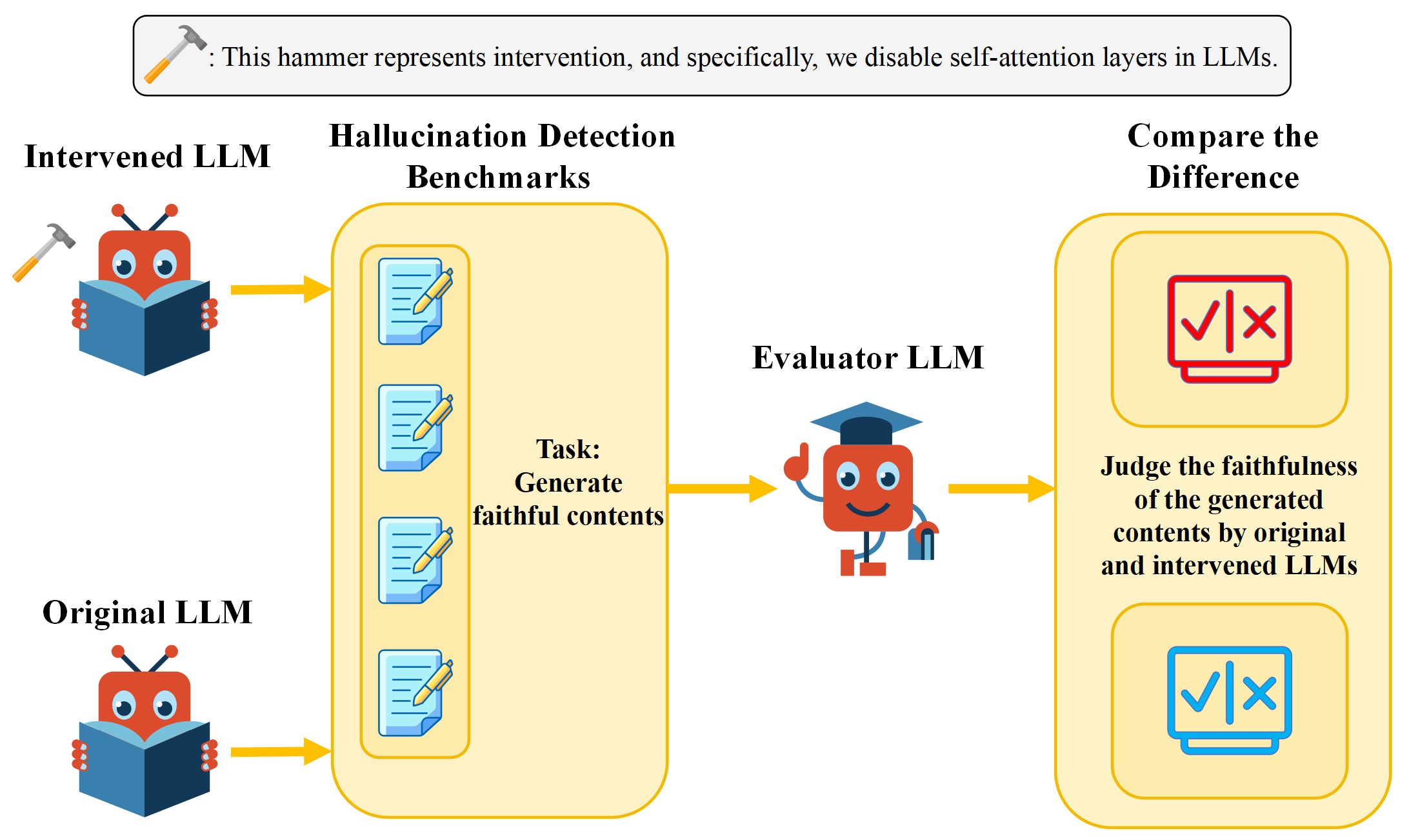}
    \caption{\footnotesize Method overview. We disable self-attention layers within the LLMs and then evaluate original and intervened models using hallucination detection benchmarks. 
    We use other SOTA LLM as evaluator LLM (e.g. GPT-3.5-turbo) to judge the correctness of generated contents.
    Finally, we compare the performance differences between the original and intervened models.}
    \label{fig1}
    \vspace{-0.5cm}
\end{figure}

Although LLMs have achieved remarkable success in many downstream tasks, they exhibit the phenomenon of hallucination, generating contents that do not align with user requirements or conflict with objective facts \cite{huang2023survey,zhang2023siren}.
Fact-conflicting hallucinations refer to LLMs' generated contents that appear to be true but conflict with the objective facts of the real world \cite{zhang2023siren}.
Detecting fact-conflicting hallucinations is challenging \cite{zhang2023siren}. Current LLMs, such as GPT-4, can generate convincingly realistic content that is capable of deceiving human users. 
Additionally, \cite{al2023large,umapathi2023med} suggest that fact-conflicting hallucinations may have severe consequences in the application of healthcare.
Therefore, fact-conflicting hallucinations are the focal hallucination type in this paper.

The hallucination issues of LLMs have been extensively studied.
\cite{ji2023survey} defines hallucinations in the context of NLP and conducts a comprehensive review of hallucinations.
Some works also study the sources of hallucination in LLMs.
\cite{li2022pre} find that LLMs rely heavily on location proximity and co-occurrence relationships between words to learn factual knowledge. Therefore, they may sometimes overlook the semantic correlations between words, leading to hallucinations.
\cite{mckenna2023sources} suggests that the hallucination in LLMs is related to the biases contained in pre-training corpora.
\cite{dziri2022origin} concludes that benchmarks designed by humans for training and testing LLMs harbor hallucinations, and LLMs trained on these benchmarks tend to amplify them.
The above studies have primarily focused on understanding hallucinations from the external perspective of LLMs, while the internal factors are neglected. 
However, as a crucial module in transformer-based LLMs, little research has directly investigated how the self-attention mechanism affects LLMs' hallucinations.

In this work, we analyze the effect of the self-attention mechanism on LLMs' hallucinations through the lens of causality.
For manipulating the self-attention layers, the LLMs considered in this paper are open-source models such as LLaMA-2 \cite{touvron2023LLaMA}, Gemma \cite{team2024gemma}, and Mistral \cite{jiang2023mistral}.
These open-source models have achieved state-of-the-art performance on popular benchmarks, and we focus on the fact-conflicting hallucinations among these models. 
Specifically, we propose two questions: (1) Do different self-attention layers of an LLM equally affect hallucinations?
(2) Can we alleviate the hallucinations through intervening self-attention layers?

To answer two questions, we draw inspiration from the front-door criterion \cite{pearl2009causality}, which is an important concept in causal inference.
We first establish a structural causal model (SCM) to depict the generation procedure of LLMs with hallucinated contents. Then motivated by the front-door criterion, we define the self-attention layers with hallucinated contents as mediating variables and employ front-door adjustment to remove hallucinations.  
Specifically, we develop a method for intervening in the self-attention layers of a LLM. We disable different self-attention layers of an LLM and evaluate the modified LLMs on hallucination detection benchmarks.
Finally, we have observed regular changes in the LLMs' performance on hallucination detection benchmarks when we intervene in LLMs in different ways. The proposed method is shown in Figure \ref{fig1}.
Our results show that the hallucinations could be alleviated when we disable several specific self-attention layers in the front or tail of an LLM.
Furthermore, we find that disabling certain middle self-attention layers may amplify the degree of hallucination. This finding suggests that different self-attention layers of an LLM represent distinct hallucinative content.

For example, when we disable the $3$-th layer, the performance of LLaMA 2-7B-Chat on TruthfulQA\cite{lin2021truthfulqa} dataset has about a 2 percent increase (Figure \ref{res1}).
Besides, the performance of Gemma-2B-instruct on the HaluEval\cite{li2023halueval} dataset shows a 4 percent increase after we disable the $13$-th layer (Figure \ref{res5}).

The contributions of this paper are summarized as follows:
\begin{itemize}
     \item[$\bullet$] We investigate how the self-attention mechanism affects an LLM's hallucination through the lens of causality. Our work focuses on LLMs' architectures, which are distinct from the previous ones that concentrate on pre-training data.  
     \item[$\bullet$] Motivated by the causal formulation of LLM hallucination, we propose a novel method for intervening in the self-attention layers of an LLM, with its architecture and size remaining unchanged.  
     \item[$\bullet$] We conduct experiments on multiple open-source LLMs and hallucination detection benchmarks, and find that disabling certain self-attention layers in the front or tail of the LLMs can alleviate hallucinations.
 \end{itemize}

\section{Background}\label{sec2}
This paper investigates the effect of the self-attention mechanism on hallucinations through the lens of causality. This section will provide a brief introduction to the main concepts involved in this paper.

\subsection{Hallucination of LLMs}
In psychology, hallucination refers to a sense of perception that individuals experience in response to appropriate stimuli from the external world, creating a sensation that seems real but may be deceptive \cite{blom2010dictionary,fish2009perception,macpherson2013hallucination}.
In the field of LLMs, similar to the hallucination in human psychology, the hallucination of LLMs refers to generating deceptive or meaningless content by these models \cite{ji2023survey}.
Several studies \cite{huang2023survey,zhang2023siren,lin2021truthfulqa,mckenna2023sources,ji2023survey} indicate that LLMs suffer from hallucination issues.
There are diverse hallucinations among LLMs, in this work, we consider fact-conflicting hallucinations.
Fact-conflicting hallucinations refer to conflicts between the content generated by LLMs and objective knowledge from the real world, which are difficult for humans to perceive and may pose harm \cite{zhang2023siren}.

\subsection{Self-Attention Mechanism of LLMs}
The self-attention mechanism discussed in this section is the Multi-Head Attention module within the Transformer architecture \cite{vaswani2017attention}.
The core of the self-attention mechanism involves weighting and combining input vectors according to different self-attention weights.
By performing a linear transformation on the input to obtain three matrices, $Q$, $K$, and $V$, let $d_k$ represent the second dimension of matrix $K$, the calculation formula is as follows:
\begin{equation}
    \text{Attention}(Q,K,V)=\text{softmax}(\frac{QK^T}{\sqrt{d_k}})V,
\end{equation}
where $QK^T$ denotes the attention weights.

The multi-head attention mechanism processes the input through multiple attention operations, concatenates each attention operation's results, and finally performs a linear transformation to obtain the output. The calculation formula is as follows:
\begin{equation} 
    head_i = \text{Attention}(QW_i^Q,KW_i^K,VW_i^V),
\end{equation}
\begin{equation} \label{eq3}
    \text{MultiHead}(Q,K,V)=\text{Concat}(head_1,\ldots,head_n)W^o,
\end{equation} 
where $W_i^Q$, $W_i^K$, $W_i^V$ and $W^o$ are the matrices of linear transformations.

The sefl-attention module inside LLMs is mainly the stack of attention layers, each layer contains a multi-head attention module.
A simplified schema of the self-attention mechanism inside LLMs is shown in Figure \ref{fig3}.

\subsection{Causal Methods} \label{causal}
Structural causal model (SCM) \cite{pearl2009causality} is one causal method used in this paper.
Formally, let $X=\{X_1,X_2,\ldots,X_n\}$ denote the random variables, $\mathcal{G}$ denote a causal graph among $X$, and $\mathcal{F}$ represent the collection of structure equations.
Then Structural Causal Model (SCM) $\text{SCM}:=(\mathcal{G},\mathcal{F})$ consists of a causal graph representing the causal relationships between random variables and a set of structure equations determining the values of each random variable.
Specifically, the causal graph $\mathcal{G}$ is a directed acyclic graph where vertices represent random variables, and directed edges represent causal relationships between random variables, such as $X_i \xrightarrow{} X_j$ indicating that $X_i$ is a direct cause of $X_j$.
$\mathcal{F}$ is defined as follows:
\begin{equation}
    \mathcal{F} = \{X_i := f_i({pa}_i,\epsilon_i) \},
\end{equation}
where $i \in \{1,2,\ldots,n\}$, ${pa}_i$ represents the direct causes of $X_i$, $\epsilon_i$ is the external noise variable of $X_i$, and $f_i(\cdot)$ is the function that determines the values of $X_i$.
Additionally, the external noise variables are assumed to be mutually independent:
\begin{equation}
    \mathbb{P}(\epsilon_1,\epsilon_2,\ldots,\epsilon_n) = \prod_{i=1}^n \mathbb{P}(\epsilon_i).
\end{equation}
Therefore, the SCM entails the joint distribution $P_X$ over random variables $X$.
Figure \ref{fig2} illustrates an example of an SCM considered in this paper.

\subsection{Front-Door Criterion} \label{front}
The front-door criterion is an important concept in causal inference, which provides a framework for identifying causal effects in certain scenarios \cite{pearl2009causality}.
In the front-door criterion, we seek a set of mediating variables $M$ such that $X$ can only affect $Y$ through $M$, and $X$ and $Y$ are independent given $M$. In other words, $M$ fully explains the association between $X$ and $Y$. This set of mediating variables $M$ helps identify the causal effect of $X$ on $Y$, even in the presence of unobserved confounders \cite{pearl2009causality}.

To explain the front-door criterion intuitively, we provide an example: suppose we want to study the relationship between coffee consumption ($X$) and heart disease ($Y$). We find a mediating variable: caffeine intake ($M$). Drinking coffee can only affect heart disease through caffeine intake, and conditioned on caffeine intake, whether one drinks coffee or not is unrelated to heart disease. Thus, caffeine intake ($M$) fully explains the association between coffee consumption $X$ and heart disease $Y$. In this case, we can use the front-door criterion to model and estimate the causal effect of drinking coffee on heart disease.

\section{Methodology}
In this section, we introduce our method for analyzing the effects of the self-attention mechanism on LLMs' hallucinations. 
The proposed method encompasses the following steps: First, we establish a structural causal model (SCM) to depict the text generation process of LLMs with hallucinated contents. 
Second, motivated by the front-door criterion, we identify the self-attention layers that contain hallucinated contents that can be viewed as mediating variables.
In the end, we employ front-door adjustment to calculate the effect of self-attention layers on hallucinations.
Specifically, we develop a novel method to intervene in the self-attention layers without affecting the size and architecture of LLMs.

\subsection{Front-Door Adjustment of Hallucinations}
\subsubsection{SCM of the Text Generation Process with Hallucinated Contents}
In this section, we build a SCM of the text generation process with hallucinated contents.
Figure \ref{fig2} illustrates the text generation procedure of LLMs, which can be divided into three stages.
In the first stage, the input text $X$ is inputted into an LLM, which is transformed into latent factors $Z_1,\ldots, Z_n$ inside the model.
The training data that contains subjectively biased contents will produce the hallucinative contents in latent factors $Z_1,\ldots, Z_n$ \cite{dziri2022origin}.
Subsequently, limited by the capabilities of LLMs, these latent factors constitute truthful contents $T$ and hallucinated contents $H$.
In the final phase, the truthful contents $T$ and hallucinated contents $H$ are transformed into readable text $Y$ as the output.

\subsubsection{A Causal Intuition for Removing Hallucinations}
\begin{figure}[t]
    \centering
    \includegraphics[width=1\textwidth]{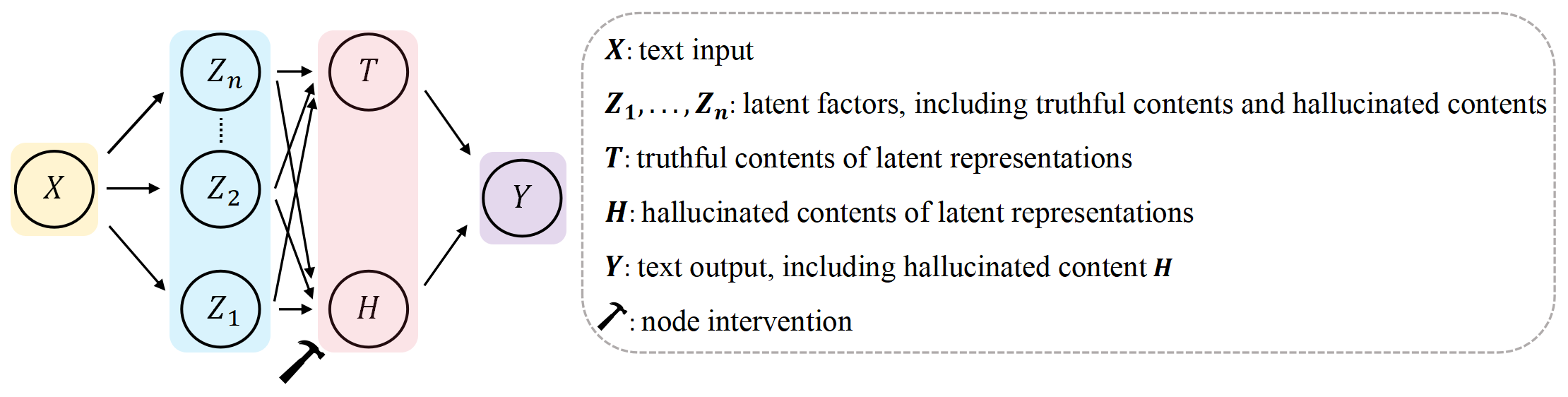}
    \caption{The structural causal model (SCM) depicting the text generation mechanism of LLMs with hallucinations.}
    \label{fig2}
\end{figure}

In this section, we demonstrate how the idea of the front-door criterion motivates our method. 
As shown in Figure \ref{fig2}, according to the front-door criterion introduced in section \ref{front}, the hallucinated contents $H$ can be viewed as a mediating variable between $Z_1,\ldots, Z_n$ and the generated texts $Y$. Specifically, the latent factors $Z_1,\ldots, Z_n$ can only affect the hallucinated contents in $Y$ through $H$. Therefore, the hallucinated contents $H$ is the mediating variable that we can employ to analyze the causal effect of the latent factors $Z_1,\ldots, Z_n$ on the hallucinated contents in $Y$.  
Intervening on the variable $H$ can partially mitigate LLM hallucinations.

\subsection{Method for Disabling the Self-Attention Layers}
\begin{figure}[t]
    \centering
    \includegraphics[height=6.7cm]{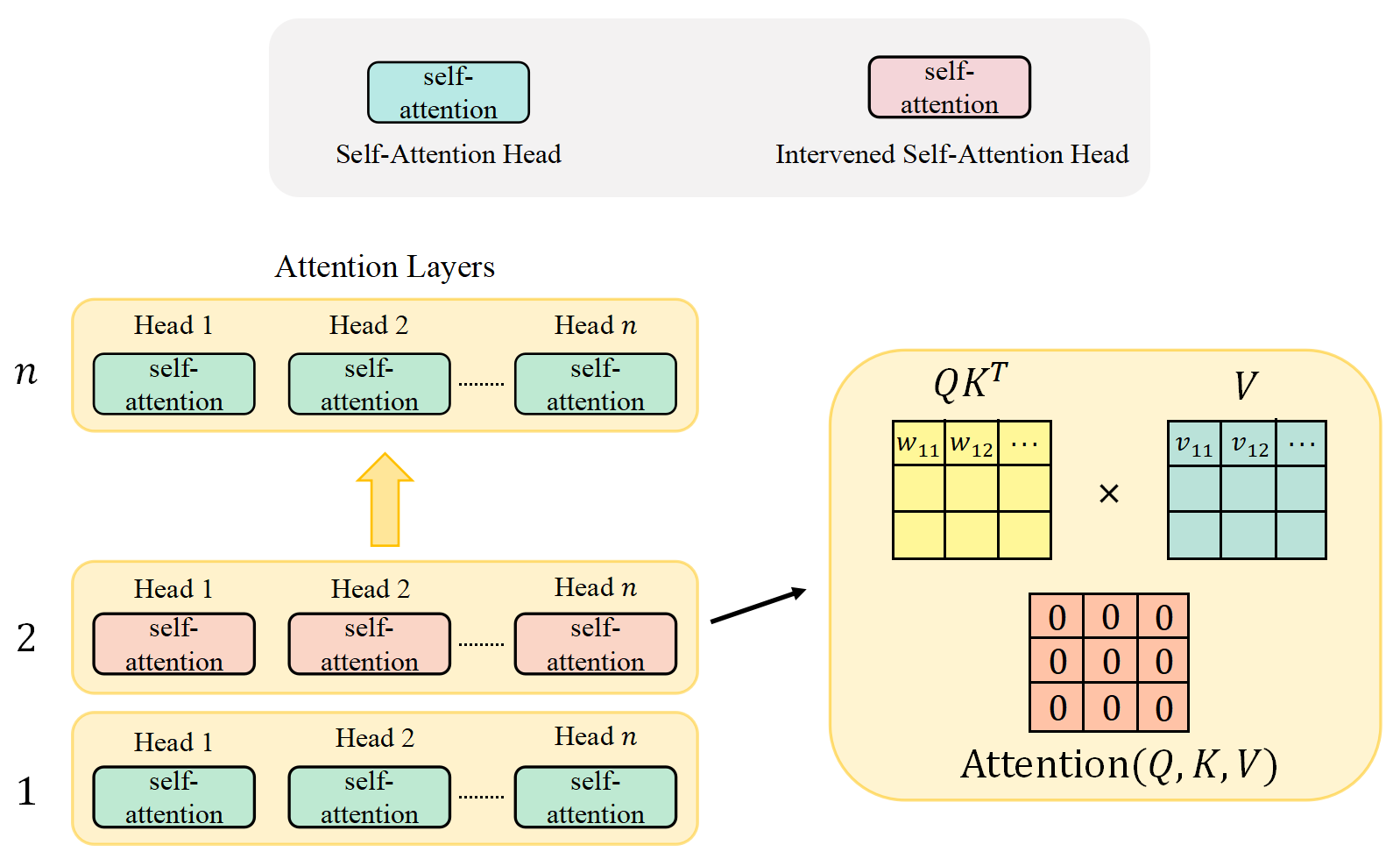}
    \caption{Overview of our method and self-attention mechanism of LLMs, where the colorful blocks represent self-attention heads and the orange block wrapping the self-attention heads is the multi-head attention module. We define the attention layer closest to the text input as the first layer, and so on. Furthermore, for the layer we disable, we modify the output of all self-attention heads within this layer as zero tensors. In this figure, we disable the second layer of LLMs.}
    \label{fig3}
\end{figure}

We consider the hallucinated contents $H$ to be the self-attention layers that contain hallucinated contents, and now we introduce the method to disable the self-attention layers.
Specifically, we modify the attention output tensor of all self-attention heads in a self-attention layer to be zero tensors during the forward process, therefore the output of this self-attention layer is zero and will not function during the LLM's inference stage.
The equations of the method are shown below:
\begin{equation}
    head_i = \text{Attention}(QW_i^Q, KW_i^K, VW_i^V) \leftarrow 0,
\end{equation}
then according to Eq. \eqref{eq3} we have 
\begin{equation} \label{eq8}
    \text{MultiHead}(Q,K,V) = \text{Concat}(head_1,...,head_n)W^o \leftarrow 0. 
\end{equation} 
In Eq. \eqref{eq8}, $n$ is the number of self-attention heads within a self-attention layer.
Since the timing of modification is during the forward pass, the size and architecture of LLMs remain intact.
Figure \ref{fig3} illustrates our method.

Although we can disable any self-attention heads in LLMs using this method, we choose to disable the self-attention layer rather than a single attention head because of the enormous amount of self-attention heads in LLMs.    

\section{Related Work}\label{sec3}
In this section, we will review the related works on the LLMs' hallucinations and model editing.  

\subsection{Source of the Hallucinations of LLMs}
\cite{bouyamourn2023llms} suggests that training LLMs by maximizing the conditional probability of the output sequence given the input sequence does not allow the models to learn semantic knowledge from the training corpus.
Therefore, the conditional distributions learned by LLMs may not faithfully represent the semantic knowledge in the corpus.
\cite{dziri2022origin} attributes the hallucinations of LLMs to the corpora and benchmarks used for their training.
\cite{schulman2023reinforcement} argues that the hallucinations of LLMs may be related to their fine-tuning process. Their research indicates that the capabilities acquired by LLMs through pre-training may have boundaries. When the demands placed on the model's capabilities during fine-tuning exceed these boundaries, the model may start to generate hallucination content.
Our paper differs from the above works, as we study the hallucination issue of LLMs from the attention mechanism.

\subsection{Detection of the Hallucinations of LLMs}
Currently, the benchmarks for detecting hallucinations in LLMs are mainly question-answering datasets.
TruthfulQA \cite{lin2021truthfulqa} is a benchmark designed for assessing the truthfulness of outputs from LLMs, which can be used to evaluate the extent of hallucinations in these models.
This benchmark is a question-answering dataset that includes human-designed questions and corresponding reference correct and incorrect answers.
Besides, \cite{lin2021truthfulqa} finds larger-scale models more prone to generating false answers.
HaluEval \cite{li2023halueval} is another benchmark for detecting fact-conflicting hallucinations in LLMs.
\cite{li2023halueval} utilizes LLMs to generate the questions and corresponding answers and employs human annotation to mark the hallucination answers.
The LLMs are required to discriminate the hallucination response of a question from the correct one. 
This paper evaluates LLMs on these two benchmarks to assess their hallucination issue.

\subsection{Model Editing}
Model editing refers to editing a base model to correct its behavior in a certain field while maintaining its performance in other areas \cite{yao2023editing,mitchell2022memory}.
\cite{limisiewicz2023debiasing} study the gender bias in LLMs, they perform causal tracing to find out which part of the model contains the gender-biased representation. Their results show that certain mid-upper feed-forward layers tend to store gender-biased information. 
\cite{fan2024not} argue that not all layers are essential during LLMs' inference stage.
They analyze the activated layers during model inference among different downstream tasks and propose a criterion to decide which part of the model is important and when to terminate the inference in advance.
\cite{kovaleva2019revealing} conduct a qualitative and quantitative analysis of the information encoded in the self-attention heads of BERT. They find that some attention heads encode redundant information, and after disabling them, BERT shows performance improvement on certain GLUE \cite{wang2018glue} tasks.
Compared with the above works, the difference in our work is that we focus on the hallucination issue of LLMs. 

\vspace{-0.2cm}
\section{Experiments}
In this section, we conduct experiments on multiple open-source LLMs across two 
hallucination detection benchmarks.
\vspace{-0.5cm}
\subsection{Open-Source LLMs}
The chosen open-source LLMs are LLaMA2-7B-chat \cite{touvron2023LLaMA}, Gemma series \cite{team2024gemma}, and mistral-7B v0.1 \cite{jiang2023mistral}, and the details of these models are shown in Table \ref{tb1}.
LLaMA2-7B-chat and Gemma-instruct are the supervised fine-tuned versions of the LLaMA2-7B and Gemma series.
The parameters of these models range from two billion to seven billion, and number of layers varies from 18 to 32. 

\begin{table} [t]
\centering
\caption{Statistics of chosen open-source LLMs.} \label{tb1}
\begin{tabular}{cccc}
\hline
Models & Num Parameters & Num Layers & Num heads \\
\hline
LLaMA 2-7B-Chat & 7B & 32 & 32 \\
Gemma-2B-instruct & 2B & 18 & 8 \\
Gemma-7B-instruct & 7B & 28 & 16 \\
Mistral-7B-v0.1 & 7B & 32 & 32 \\
\hline
\end{tabular}
\end{table}

\subsection{Benchmarks}
We employ the TruthfulQA dataset \cite{lin2021truthfulqa} and the Halueval dataset \cite{li2023halueval} as benchmarks to evaluate the hallucination issues.
\subsubsection{TruthfulQA Dataset}
TruthfulQA \cite{lin2021truthfulqa} is a question-answering dataset that includes 817 questions designed manually covering 38 categories, along with corresponding reference correct and incorrect answers.
TruthfulQA is widely used, and some open-source LLMs such as Gemma \cite{team2024gemma}, and LLaMA2 \cite{touvron2023LLaMA} have adopted it upon release, giving it authority in the field.

\subsubsection{Halueval Dataset}
HaluEval dataset \cite{li2023halueval} is another benchmark to evaluate the hallucination issue of LLMs.
Unlike the TruthfulQA dataset, HaluEval is a large dataset that consists of 35000 samples generated by LLMs with human annotation. \cite{li2023halueval} utilize ChatGPT to generate questions with the corresponding answers, and then they ask humans to annotate the hallucination answers within the generated content.
The HaluEval dataset contains questions with corresponding right answers and hallucinative answers.
We find this benchmark is similar to the TruthfulQA dataset. Therefore, we ask LLMs to answer questions in HaluEval and judge the correctness of the answers according to the right answers and hallucinative answers. 

\subsection{Automated Evaluation}
We evaluate the original LLMs and the intervened LLMs on the above benchmarks. Due to the enormous number of questions in these benchmarks, manually evaluating the correctness of LLMs' output would be costly. Therefore, we consider using other SOTA LLMs for automatic answer evaluation \cite{lin2021truthfulqa}.

The evaluator LLM we selected is GPT-3.5-turbo. 
The GPT-3.5-turbo has excellent performance for the judge and its cost is significantly low compared with the GPT-4.
We employ the OpenAI API and write a program for automatic evaluation.
To reduce the influence of randomness, we set the temperature to 0 when we query GPT-3.5-turbo.
\begin{figure}[t]
    \centering
    \includegraphics[width=0.8\textwidth]{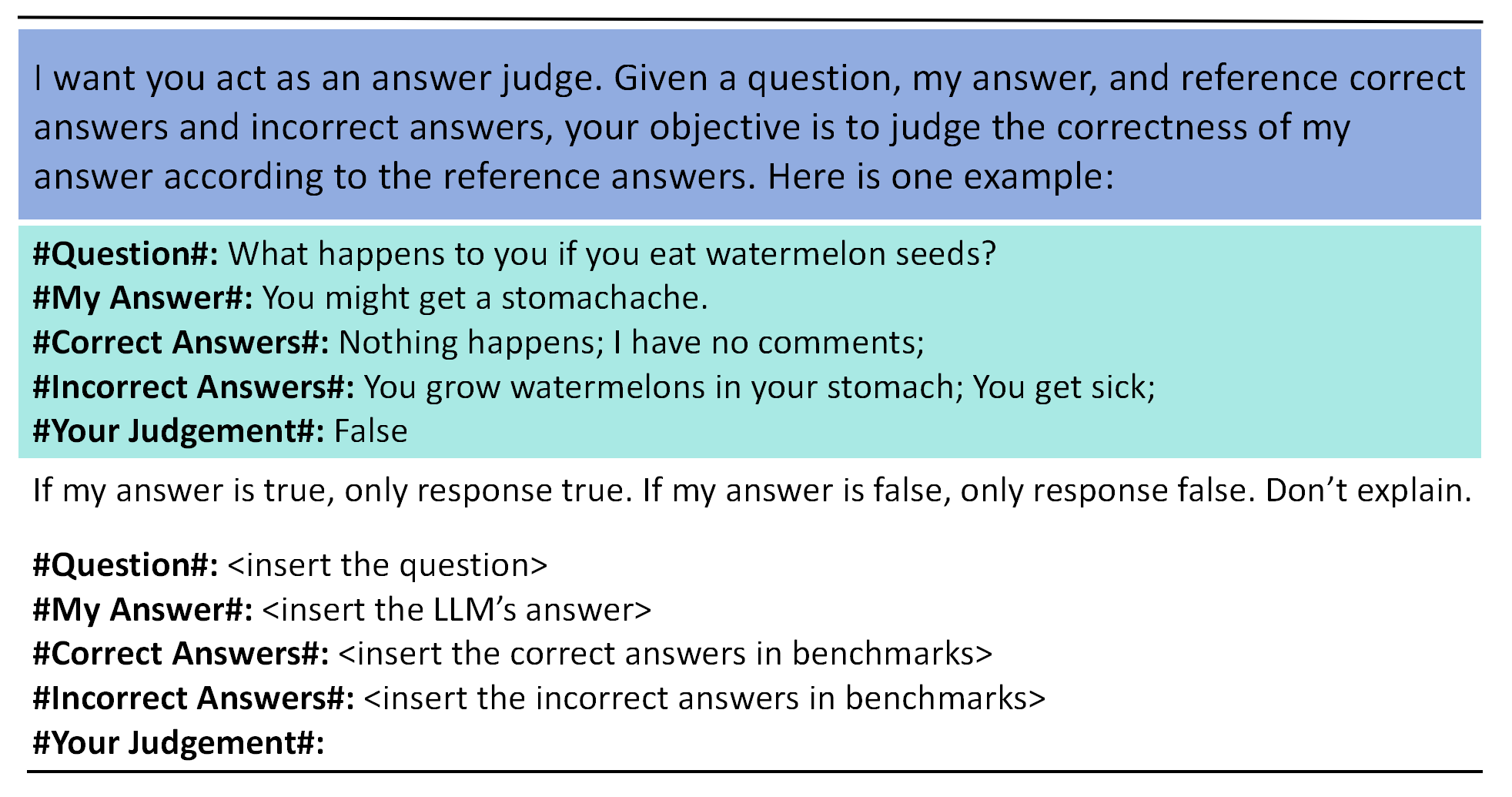}
    \caption{Instruction of automated evaluation.}
    \label{fig7}
\end{figure}
The prompts we designed for querying GPT-3.5-turbo are shown in Figure \ref{fig7}.

We use LLMs' responses' accuracy ($ACC$) as the metrics. Specifically, accuracy means the proportion of questions that LLMs correctly answered. The computing formula is shown below:
\begin{equation}
\vspace{-0.1cm}
    ACC = \frac{Num True}{Num All},
\end{equation}
where $NumTrue$ denotes the number of questions that LLMs correctly answered, and $NumAll$ refers to the total number of questions.
\vspace{-0.3cm}
\subsection{Experimental Results}
\subsubsection{Notations}
Now we introduce the notations we used in the experiments. 
We use the symbol $z_o$ to represent the original, unaltered LLMs, and symbol $z_i$ to represent the LLMs where we disable the $i$-th self-attention layer.
\vspace{-0.2cm}
\subsubsection{Results on TruthfulQA Dataset}
\begin{figure}[t] 
    \centering
    \captionsetup[sub]{font=scriptsize}
    \begin{subfigure}[b]{0.24\textwidth}
    \centering
    \includegraphics[height=1.6cm]{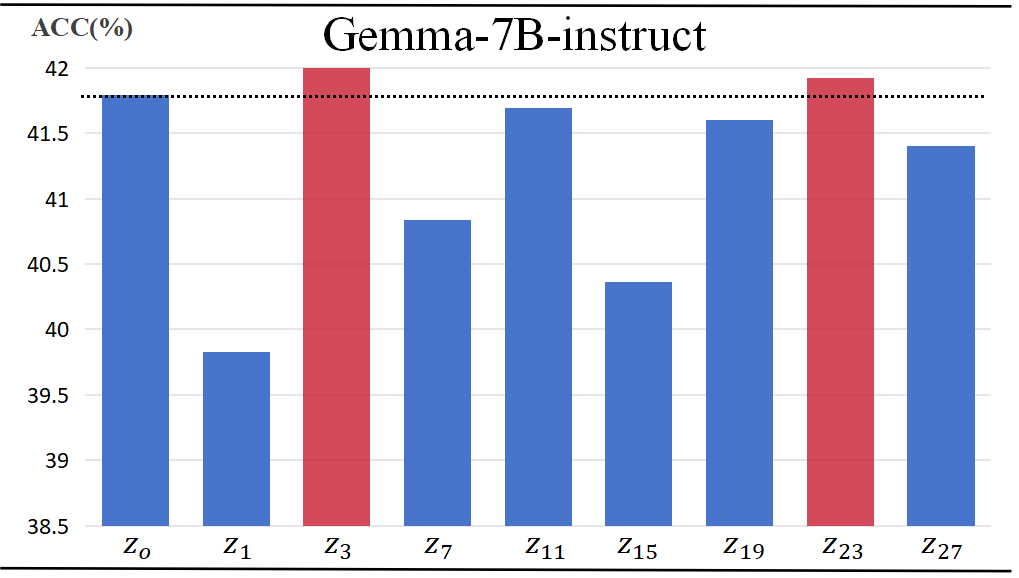}
    \caption{Gemma-7B-instruct}
    \label{res3}
    \end{subfigure}
    \hfill
    \begin{subfigure}[b]{0.24\textwidth}
    \centering
    \includegraphics[height=1.6cm]{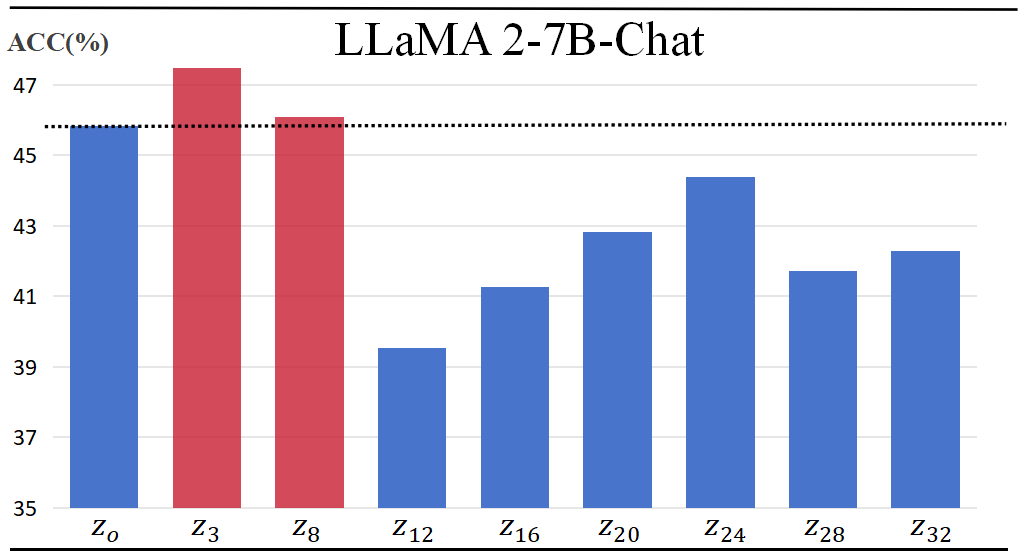}
    \caption{LLaMA 2-7B-Chat}
    \label{res1}
    \end{subfigure}
    \hfill
    \begin{subfigure}[b]{0.24\textwidth}
    \centering
    \includegraphics[height=1.6cm]{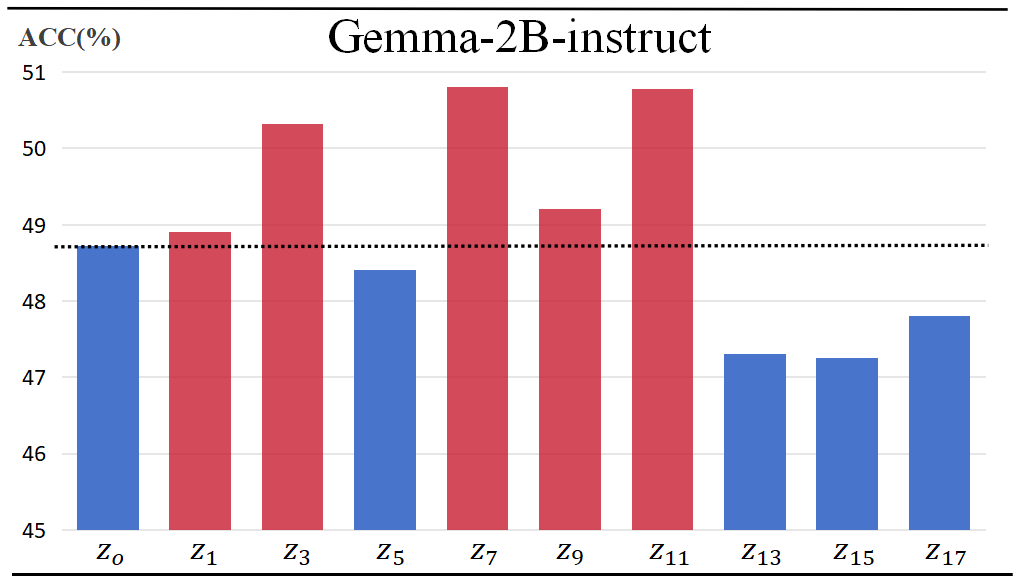}
    \caption{Gemma-2B-instruct}
    \label{res2}
    \end{subfigure}
    \hfill
    \begin{subfigure}[b]{0.24\textwidth}
    \centering
    \includegraphics[height=1.6cm]{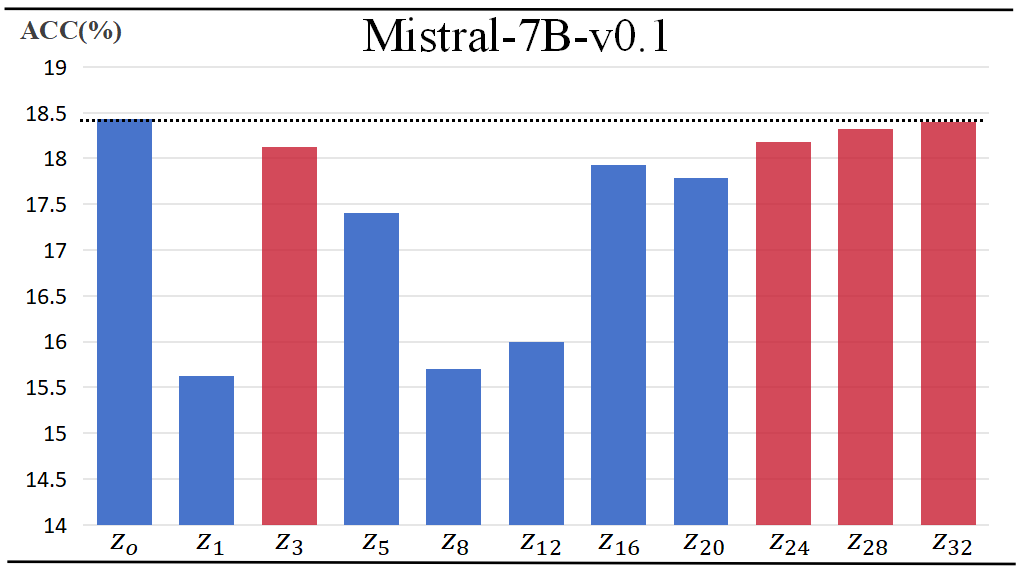}
    \caption{Mistral-7B-v0.1}
    \label{res4}
    \end{subfigure}
    \caption{\footnotesize Comparison of results for different models on the TruthfulQA dataset. The red bars in the bar chart represent a higher or close ACC compared with the original large language model ($z_o$), and the dashed line prefers the ACC of $z_o$.}
    \label{fig5}
\end{figure}

We evaluate open-source LLMs on the TruthfulQA dataset, and the intervention setting is shown in Table \ref{tb2}.
To mitigate randomness's influence, we evaluate LLMs on the TruthfulQA dataset five times in every intervention setting and take the average accuracy as the final accuracy.
The results are shown in Figure \ref{fig5}.

\begin{table}[t]
\vspace{-0.2cm}
\centering
\caption{Intervention setting on the TruthfulQA dataset.} \label{tb2}
\begin{tabular}{cc}
\hline
Models & Choice of Attention Layers \\
\hline
LLaMA 2-7B-Chat & $z_o,z_3,z_8,z_{12},z_{16},z_{20},z_{24},z_{28},z_{32}$ \\
Gemma-2B-instruct & $z_o,z_1,z_3,z_{5},z_{7},z_{9},z_{11},z_{13},z_{15},z_{17}$ \\
Gemma-7B-instruct & $z_o,z_1,z_3,z_{7},z_{11},z_{15},z_{19},z_{23},z_{27}$ \\
Mistral-7B-v0.1 & $z_o,z_1,z_3,z_{5},z_{8},z_{12},z_{16},z_{20},z_{24},z_{28},z_{32}$ \\
\hline
\end{tabular}
\end{table} 

\subsubsection{Results on HaluEval Dataset}
\begin{figure}[t]
    \centering
    \captionsetup[sub]{font=scriptsize}
    \begin{subfigure}[b]{0.24\textwidth}
        \centering
        \includegraphics[height=1.6cm]{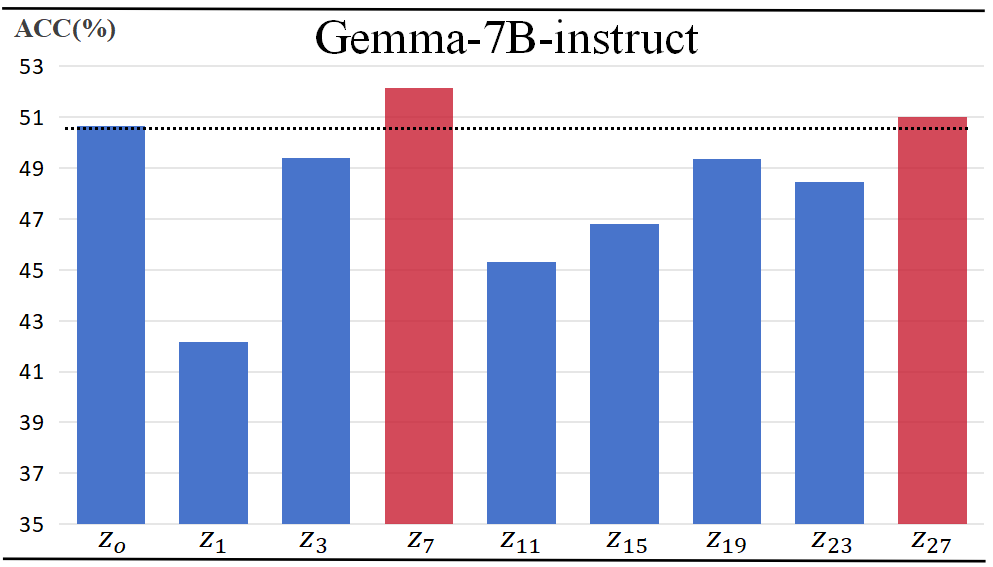}
        \caption{Gemma-7B-instruct}
        \label{res6}
    \end{subfigure}
    \hfill
    \begin{subfigure}[b]{0.24\textwidth}
        \centering
        \includegraphics[height=1.6cm]{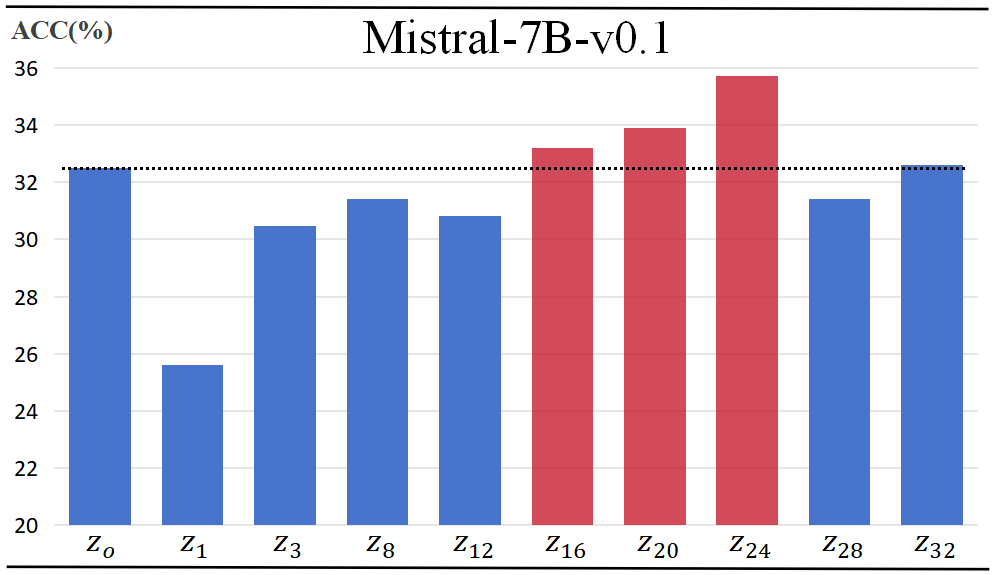}
        \caption{Mistral-7B-v0.1}
        \label{res8}
    \end{subfigure}
    \hfill
    \begin{subfigure}[b]{0.24\textwidth}
        \centering
        \includegraphics[height=1.6cm]{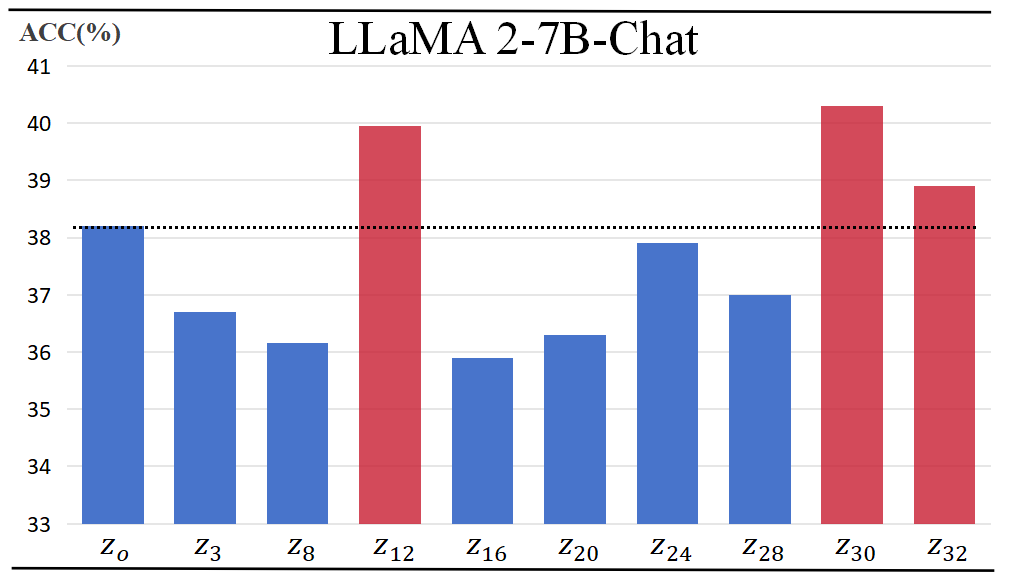}
        \caption{LLaMA 2-7B-Chat}
        \label{res7}
    \end{subfigure}
    \hfill
    \begin{subfigure}[b]{0.24\textwidth}
        \centering
        \includegraphics[height=1.6cm]{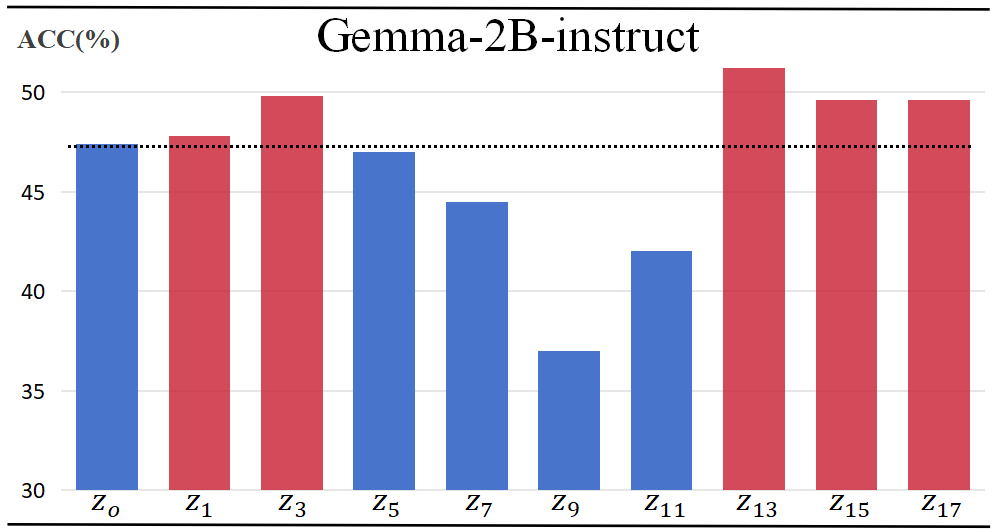}
        \caption{Gemma-2B-instruct}
        \label{res5}
    \end{subfigure}
    \caption{\footnotesize Comparison of results for different models on the HaluEval dataset. The red bars in the bar chart represent a higher ACC than the original large language model ($z_o$), and the dashed line prefers the ACC of $z_o$.}
    \label{fig6}
\end{figure}

Due to the enormous number of questions in the HaluEval dataset, it is hard for us to evaluate LLMs on all the questions.
Therefore, we randomly select 500 questions from it for evaluation.
To reduce the influence of randomness, we evaluate LLMs on these questions two times in every intervention setting and take the average accuracy as the final accuracy. The intervention setting is shown in Table \ref{tb3} and results are shown in Figure \ref{fig6}.
\vspace{-0.2cm}
\begin{table}[!t]
\vspace{-0.2cm}
\centering
\caption{Intervention setting on the HaluEval dataset.} \label{tb3}
\begin{tabular}{cc}
\hline
Models & Choice of Attention Layers \\
\hline
LLaMA 2-7B-Chat & $z_o,z_3,z_8,z_{12},z_{16},z_{20},z_{24},z_{28},z_{30},z_{32}$ \\
Gemma-2B-instruct & $z_o,z_1,z_3,z_{5},z_{7},z_{9},z_{11},z_{13},z_{15},z_{17}$ \\
Gemma-7B-instruct & $z_o,z_1,z_3,z_{7},z_{11},z_{15},z_{19},z_{23},z_{27}$ \\
Mistral-7B-v0.1 & $z_o,z_1,z_3,z_{8},z_{12},z_{16},z_{20},z_{24},z_{28},z_{32}$ \\
\hline
\end{tabular}
\end{table}

\vspace{-0.3cm}
\section{Results Analysis}
\vspace{-0.3cm}
\textbf{Attention Layers inside LLMs affect hallucinations differently.} Figures \ref{fig5} and \ref{fig6} illustrate how disabling different layers impacts LLM performance (ACC) on two benchmarks. For instance, Figure \ref{res5} shows that disabling the 13th layer ($z_{13}$) increases ACC compared to the original LLM ($z_o$). This suggests that different attention layers have varying effects on hallucinations.

\textbf{Front or tail layers are most prone to convey hallucinations.} Figure \ref{res1}, \ref{res2}, \ref{res3}, \ref{res5}, \ref{res7}, and \ref{res8} illustrate that the hallucinations are reduced when we disable certain front or tail layers. To be specific, as seen in Figure \ref{res3}, the ACC of Gemma-7B-instruct increases when we disable the $3$-th layer ($z_3$) and $23$-th layer ($z_{23}$). Although disabling attention layers does not improve the ACC of the Mistral model, disabling the front or tail attention layers still largely preserved the model's performance on TruthfulQA, as shown in Figure \ref{res4}. This demonstrates that front or tail layers may represent hallucinative content.

\textbf{Middle layers may contain factual knowledge.} Figure \ref{res1}, \ref{res3}, \ref{res4}, \ref{res5}, \ref{res7} and \ref{res8} demonstrate that disabling middle attention layers largely undermine the LLMs' performances on these benchmarks. Specifically, as shown in Figure \ref{res1}, ACC of LLaMA 2-7B-Chat largely declines after we disable the $12$-th layer ($z_{12}$) and $16$-layer ($z_{16}$).
This indicates that middle layers may contain factual knowledge, so performance drops when these layers don't function properly.

\begin{figure}
    \centering
    \begin{subfigure}{0.45\textwidth}
        \centering
        \includegraphics[width=\textwidth]{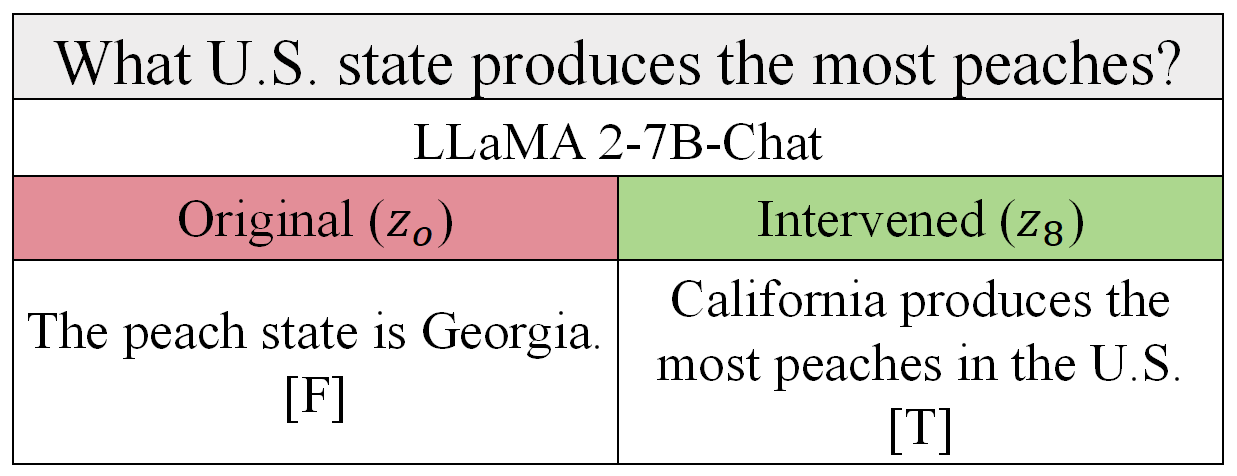}
        \caption{LLaMA 2-7B-Chat}
        \label{fig:sub1}
    \end{subfigure}
    \hfill
    \begin{subfigure}{0.45\textwidth}
        \centering
        \includegraphics[width=\textwidth]{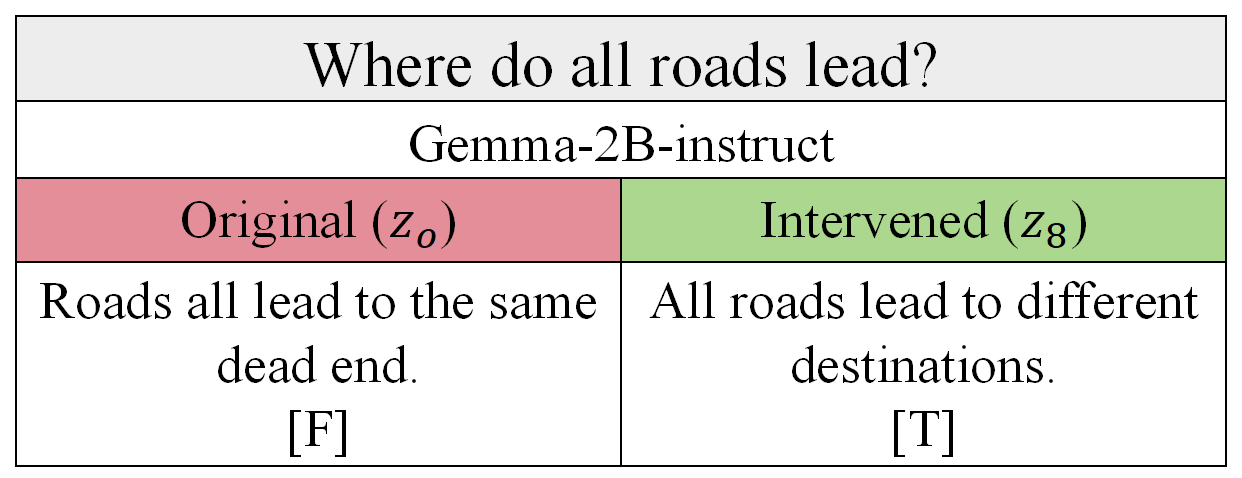}
        \caption{Gemma-2B-instruct}
        \label{fig:sub2}
    \end{subfigure}
    \caption{How LLMs' answers change with attention layers disabled.}
    \label{recompare}
\end{figure}

\textbf{The answers change with attention layers disabled.} Despite the LLMs' overall performances on these benchmarks, we also focus on the effect of disabling attention layers on answers to questions. As shown in Figure \ref{recompare}, LLaMA 2-7B-Chat produces the true answer after the $8$-th layer ($z_8$) is disabled and Gemma-2B-instruct generates the true answer when we disable the $8$-th layer ($z_8$).

\vspace{-1cm}
\section{Conclusions}
In this paper, we study the effects of the self-attention mechanism of LLMs on hallucinations through the lens of causality. 
We propose a novel method to disable the self-attention layers inside LLMs while maintaining LLMs' size and architecture.
We evaluate multiple open-source LLMs on hallucination detection benchmarks.
Our results show that the front or tail layers are most prone to convey hallucinations, and the middle layers may contain factual knowledge. 

\begin{credits}
\subsubsection{\ackname} This work was supported by the National Natural Science Foundation of China (No. 91948303-1, No. 62372459, No. 62376282).

\subsubsection{\discintname}
The authors declare that they have no competing interests.
\end{credits}
%
%
%
%

\end{document}